\documentclass[letterpaper, 10pt, conference]{ieeeconf}
\IEEEoverridecommandlockouts
\usepackage{cite}
\usepackage{amsmath,amssymb,amsfonts}
\usepackage{algorithmic}
\usepackage{graphicx}
\usepackage{lipsum}
\usepackage{textcomp}
\usepackage{xcolor}
\usepackage{todonotes}
\def\BibTeX{{\rm B\kern-.05em{\sc i\kern-.025em b}\kern-.08em
    T\kern-.1667em\lower.7ex\hbox{E}\kern-.125emX}}
\begin{document}


\binoppenalty=10000 
\relpenalty=10000

\title{\LARGE \bf Dynamic Modeling, Gait Synthesis, and Control of a Novel Subsurface Bore Propagator}

\author{Lina van Br\"ugge, Shruti Kotpalliwar, Akshit Saradagi, Anton Koval, George Nikolakopoulos}

\maketitle

\begin{abstract}
In this article, we present dynamic modeling, gait synthesis, and feedback control design for a modular novel subsurface robot, designed for human-free subsurface exploration and excavation. The subsurface propagator design is based on two major aspects: 1) anchor and propel movement like an earthworm and 2) excavation similar to tunnel boring machines. This design is decoupled into five separate modules: one drill head to excavate and create cavity for propagation, two modules to anchor the robot, and two modules to enable propagation of the body. In order to design a controller for each of the modules, dynamic models using the Euler-Lagrange framework are developed. These mathematical models are used as a baseline to design controlled decoupled operation of the different joint movements. The operation of robotic assembly is constructed via a centralized state machine for gait synthesis with integration of the designed feedback controller. The controllers are tested on the real robot geometry to aid sim-to-real integration: A physics-based Unity simulation using a CAD model of the robot and integration of the trained controller via ROS verifies the performance of the robot. The experimental results demonstrate that the proposed design, controllers and the gait synthesis strategy together are capable of anchoring the robot in place and creating an total advancement of 30\,mm into the soil after completing 3 gait cycles.
\end{abstract}
%
\section{Introduction}
%
Autonomous exploration of underground environments introduces several challenges, such as lack of hollow space for movement, a feature-poor environment, and limited sensing capabilities for the deployment of automated robotic systems. Thus, subterranean operations demand robot designs capable of creating space and moving in a confined tunnel-like environments with restricted sensing capabilities~\cite{yoshida_excavation_2023}. 
The central element of underground excavation is the design of a mechanism to excavate and propagate while generating enough force to counteract forces from the environment \cite{dorgan_fundamentals_2023}. Therefore, these designs often incorporate bio-inspiration from burrowing animals such as moles, worms, or snakes for the creation of space, anchoring and movement in confined spaces \cite{kim_development_2018, lee_development_2020, zhang_mole-inspired_2024, yoshida_excavation_2023},  \cite{das_earthworm-like_2023, omori_2009, liu_design_2024}. In \cite{kimDevelopmentMoleLikeDrilling2018}, a robot mimicking mole-like burrowing is presented, integrating excavation, propulsion and soil removal directly in one platform. Further development concentrated on different drilling bits combining anchoring and excavation in a single unit \cite{lee_concept_2019, lee_development_2020}, however, directional drilling cannot be achieved with this design. Also the robot in \cite{das_earthworm-like_2023} performs solely linear movement. However, the worm-like robot achieves burrowing as well as locomotion within a straight pipe using peristaltic soft actuators. Other mole-like robots extend their movement to planar locomotion, demonstrating trajectory following and burrowing in granular medium \cite{zhang_mole-inspired_2024} while highlighting the relation between limb speed and forward propagation.

Snake-like robots also incorporate capability to move in confined spaces using obstacle aided locomotion. The authors in \cite{mai_application_nodate}  present a snake-like robot able to anchor itself by deforming its body to press against pipe walls. This principle is also adopted on another snake-like excavation robot with drill head and fins \cite{yoshida_excavation_2023}. For this method, a significant difference in diameter between the robots and the tunnel is necessary. Other designs are therefore concentrating on worm-like motion principles: In \cite{omori_2009}, an underground explorer robot able to perform 3D-dimensional movement is presented. Its capabilities are demonstrated during vertical climbing inside a pipe and conformed dirt, showing the versatility of worm-like movement for underground tasks. Also in \cite{liu_design_2024}, worm-like locomotion is chosen for pipe crawling. The robot uses telescopic modules for anchoring and propagation and 2D-joint modules for turning. Underground excavation is also targeted in \cite{vartholomeos_modeling_2021}. Using a similar approach as in \cite{liu_design_2024}, the BADGER robot is used for shallow drilling between two excavated pits for pipe installation. Linear drilling was demonstrated in experiments while 3-dimensional trajectory following was verified in simulations.

As the general characteristics of worm-like locomotion can be used to increase the control effectivity, research focuses on the dynamics of peristaltic motion itself. In \cite{shi_continuum_2026}, the continuum dynamics of peristaltic movement are modeled enabling the analysis of the global behavior of the robot while preserving the essential mechanics of metameric movement. Different cost functions to maximize movement velocity or minimize cost of transport are established in \cite{kandhari_analysis_2021} and demonstrated on a soft robot with a compliant modular mesh. A similar robot is also used to demonstrate a stable heteroclinic channel controller which reduces up to 40\% slip during propagation compared to a best-fit controller \cite{daltorio_efficient_2013}.

Despite significant progress, gaps still persist. To emphasize, \emph{co-design} of drilling heads, anchoring/bracing subsystems, body configuration planning, and robot control under \emph{quantified} granular mechanics is rarely addressed in a unified framework. Most works optimize one layer (e.g. bit mechanics~\cite{kim_development_2018,lee_development_2020} or gait control \cite{daltorio_efficient_2013, zhou_cpg-based_2023}), assuming the other layers are fixed. Additionally, designs targeting similar applications such as the BADGER robot have not yet been tested in their complete design space (e.g. just linear propulsion \cite{omori_2009, vartholomeos_modeling_2021} or separate tests on drilling and locomotion \cite{kim_development_2018}). This clearly demonstrates the difficulties arising from the design itself and its experimentation.

\begin{figure}[h]
    \centering
    \includegraphics[width=0.45\textwidth]{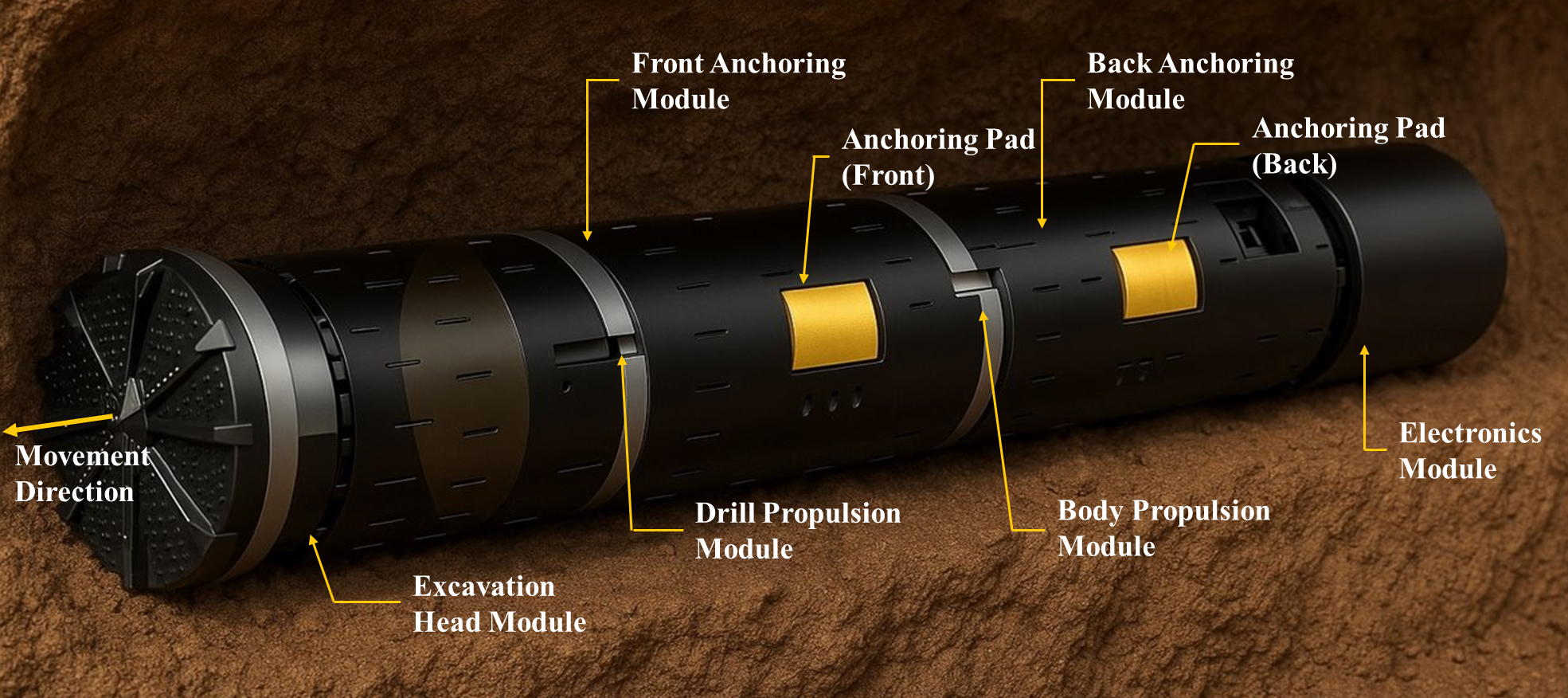}
    \caption{Concept of the bore propagator  (CAD model of the robot in a soil type sub-surface) }
    \label{fig:concept}
    \end{figure}

In this work, we consider a novel robot design inspired by the drilling mechanism of tunnel-boring machines and the locomotion strategy of an earthworm. The design concept is deconstructed into: a) two anchoring modules (AMs) (front and back) , b) two propulsion modules (PMs) (drill and body), and c) one excavation head module (EHM); see Fig. \ref{fig:concept}. Both AMs are equipped with pads that anchor the robot to the pre-drilled surface, while PMs propel the robot body forward or backward through the motors installed in these modules. The EHM at the head of the robot drills through the subsurface, and the drilled material passes through the robot's body via an auger mechanism and is expelled from the tail of the robot.
Overall, the main contribution of this article are: I) A mathematical framework for faster and easier verification of robot's operation. The framework consist of 1. The system is decoupled into a few modules and the mathematical description of each module is obtained using the Euler-Lagrange framework separately; 2. these mathematical descriptions are then used to design feedback control for each of these modules. 3. To combine the decoupled components a centralized state-machine based gait synthesis is designed. 1, 2, and 3 are obtained via a MATLAB/Simulink simulation and establish a baseline of the operation procedure of the complete robot. 1 and 2 allow faster and easier verification of the controllers in the MATLAB/Simulink based environment, and the decoupled components provide flexibility to obtain dynamics of robots with different module configurations. The controllers tuned in this framework are lifted for: contribution II) sim-to-real integration. Here, we use the CAD geometry of the robot in Unity \cite{unity} together with controllers derived from the math based models (I.1 + I.2) and gait synthesis (I.3) interfaced via ROS.  This simulation in Unity mimics the real environment condition and robot geometry while maintaining the foundational model based controller design.

\section{Design and Mathematical Modeling of the Subsurface Bore Propagator}
\label{sec:description-robot}
\begin{figure}
    \centering
    \includegraphics[width=\linewidth]{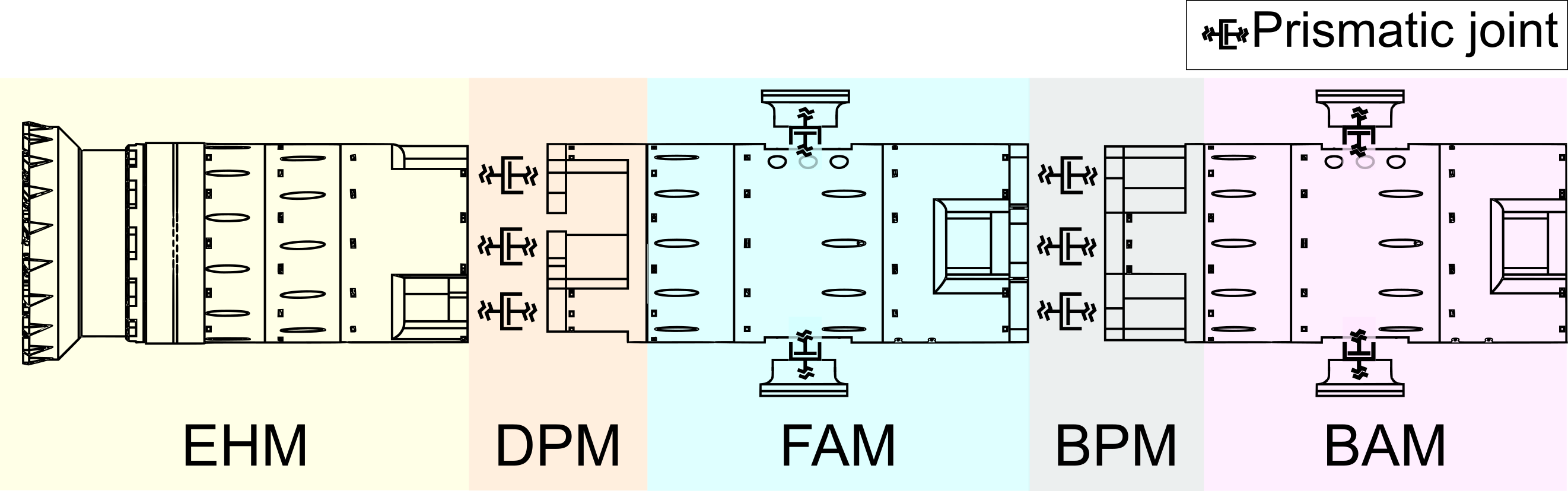}
    \caption{Robot with connecting joints}
    \label{fig:robot_joints}
\end{figure}

The robot assembly, see Fig.~\ref{fig:concept}, starts with EHM for drilling into the soil incorporating the drill to create a tunnel and the actuation unit, including the motor, gears, and bearings. The next is the Drill PM (DPM) that connects the EHM and the Front AM (FAM) through a three lead-screw-based system evenly distributed around the perimeter of the robot. When actuated simultaneously, these screws push the connected module forward. Following the DPM, is the FAM which provides counteracting forces through firm anchoring during drilling and locomotion. This mechanism is achieved using two lead-screws connected to servo motors that engage or disengage the pads and press them onto the tunnel walls. Following the FAM, are the Body PM (BPM) and the Back AM (BAM) these two mimic the operation of the DPM and the FAM respectively. The two AMs provide the strength required to counteract vibrations introduced by drilling and maintain stable position during drilling operation and propagation. In the remainder of this section, we present a mathematical model of each module and for this the lead-screw-based systems in PM and AM are depicted as prismatic joints, as shown in Fig. \ref{fig:robot_joints}. 


The equations of motion (EoM) of each module are derived separately. This is possible as the modules are generally independent of each other. However, to account for forces and moments in connecting modules, each module features additional coordinate systems (CoSys) at the start and end of each module, and wrenches are applied there.

\subsection{Dynamic Model of the Anchoring Module}
\begin{figure}[h]
    \centering
    \includegraphics[width=0.7\linewidth]{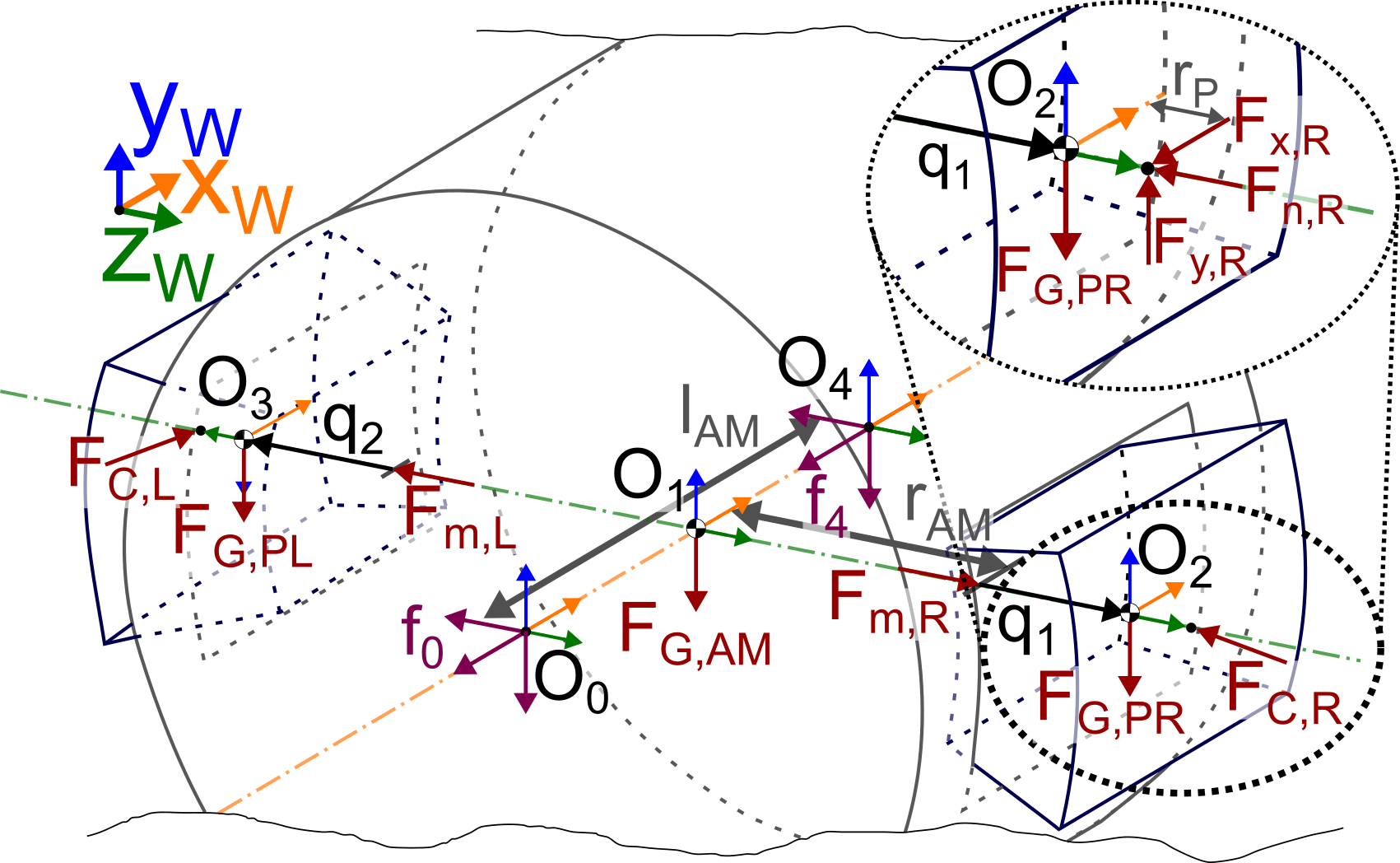}
    \caption{Coordinate systems and forces used to describe the AM}
    \label{fig:forces_am}
\end{figure}

The connection of the pads to the main body of the AM can be approximated by a prismatic joint. This results in two degree of freedoms (dofs) $q_{1,AM}\in[0,25e^{-3}]$\,mm and $q_{2,AM}\in[-25e^{-3},0]$\,mm acting on the center of mass (CoM) of the pads and therefore describing their position. The complete module can be described using 5 CoSys as shown in Fig.~\ref{fig:forces_am}. CoSys $O_0$, $O_1$ and $O_4$ are fixed to the main body describing the position of connecting modules and its CoM. CoSys $O_2$ and $O_3$ describe the position of the right and left pads, respectively. The corresponding Denavit-Hartenberg (DH) parameters are presented in Table \ref{tab:dh-am-parameter}. 

\begin{table}[h]
    \centering
    \caption{DH parameter for describing the transforms of the Anchoring Modules}
    \begin{tabular}{cccccc}
    \hline
    Link$_i$ & Link$_j$& $a_i$ & $\alpha_i$ & $d_i$ & $\theta_i$\\
    \hline
    0 & 1 & $l_{AM}/2$ & 0 & 0 & 0\\
    1 & 2 & 0 & 0 & $r_{AM}+q_1$ & 0\\
    1 & 3 & 0 & $\pi$ & -($r_{AM}+q_2$) & 0\\
    1 & 4 & $l_{AM}/2$ & 0 & 0 & 0\\
    \hline
    \end{tabular}
    \label{tab:dh-am-parameter}
\end{table}

\subsubsection{Kinematic Relations}
The transformations $T_{ref\rightarrow i}=\begin{bmatrix}R_i & p_i\\0&1\end{bmatrix}$ for the separate CoSys can be described following the classic DH scheme. For deriving the EoM, $O_1$ is selected as the reference frame. This simplifies the model and removes torques resulting from the otherwise necessary transformation $T_{i\rightarrow0}$. As the robot generally does not touch the ground during deployment, its base has to be assumed to be free-floating in space. This results in the vector of generalized velocities $\dot q=[\omega_0~v_0~\dot q_1~\dot q_2]^T$ where $\vec \omega_0\in\Re^{3\times1}$ and $\vec v_0\in\Re^{3\times1}$ denote the rotatory and translatory velocities of the base frame $O_0$. Since frames $O_0$, $O_1$ and $O_4$ are on the same rigid body, it applies $v_4 = v_1 = v_0$. In order to compute the velocities of the pads, the Jacobians $J_{2/3}$ have to be derived. This is done by transforming the vectors in spatial notation. In this notation, the motion space of the prismatic joints can be described by $S_{1/2}=[0~0~0~0~0~1]^T$.

Motions are chosen to be expressed with respect to the reference frame $O_1$. Thus, the spatial motion transform as well as the Jacobians do not include rotational elements. Using the skew operator $[p]_\times$ which is defined by $[p]_\times x=p\times x$, the spatial motion transform $X_{i\rightarrow 1}$ from the $O_i$ to the reference frame $O_1$ can be computed by
\begin{equation}
    X_{1\rightarrow i} = \begin{bmatrix}
        I_3 & 0\\-[p_i]_\times & I_3
    \end{bmatrix},
    \label{eq:spatial_motion_transform}
\end{equation}
where $p_i$ defines the translation from the reference frame $O_1$ to frame $O_i$. The velocity of the two pads can then be computed by Eq.~\eqref{eq:velocity_pads}.
\begin{equation}
\begin{aligned}
    v_{1/2} &=&J_{1/2}\dot q,~\text{with}\\
    J_2&=&[X_{1\rightarrow2}~S_1~0]\\
    J_3&=&[X_{1\rightarrow3}~0~S_2] 
\end{aligned}
\label{eq:velocity_pads}
 \end{equation}

\subsubsection{Dynamic Relations}
Eq.~\eqref{eq:eom} describes the EoM of the AM. In the following section, the separate terms are derived dependent on the forces acting upon the AM as shown in Fig.~\ref{fig:forces_am}.

\begin{equation}
    M(q)\ddot{q} + C(q,\dot q)\dot q = \tau_C(q)+ \tau_{ext}(q) + \tau_{act}(q) + G(q)
    \label{eq:eom}
\end{equation}

\paragraph{Mass Matrix}
The mass matrix is time-variant as it depends on the pad positions and therefore on $q_{1,AM}$ and $q_{2,AM}$. To fully define the mass matrix, the Jacobian for the CoM of the reference frame $O_1$ $J_1 = [I_6~0~0]$ is necessary. The total mass matrix is then assembled using
\begin{equation}
    M(q)=\sum^3_{i=1}J_i^TI_iJ_i, \text{ with } I_i=\begin{bmatrix}
        I_{i,rot}&0\\0&m_iI_3
    \end{bmatrix}.
    \label{eq:mass-matrix}
\end{equation}

\paragraph{Coriolis Term}
Due to the time-variance of the mass matrix $M(q)$, a Coriolis term has to be added to the EoM. To model the Coriolis matrix $C(q,\dot q)$, Christoffel symbols of the first kind are used \cite{spong2020} as described by
\begin{equation}
    [C(q,\dot q)]_{ij}=\sum_k \Gamma_{ijk}(q)\dot q_k
    \label{eq:coriolis_christoffel}
\end{equation}
where
\begin{equation}
    \Gamma_{ijk}=\frac{1}{2}\left( \frac{\partial M_{ij}}{\partial k}+\frac{\partial M_{ik}}{\partial j}+\frac{\partial M_{jk}}{\partial i}\right)\end{equation}.

As the mass matrix is solely dependent on $q_{1,AM}$ and $q_{2,AM}$, Eq.~\eqref{eq:coriolis_christoffel} simplifies to
\begin{equation}
    C(q,\dot q) = \frac{1}{2}\left(\frac{\partial M}{\partial q_1}\dot q_1+\frac{\partial M}{\partial q_2}\dot q_2\right).
    \label{eq:christoffeles_derivative}
\end{equation}

\paragraph{Gravity Force}
All forces have to be transformed with respect to the reference frame $O_1$. This is done using the spatial force transform $X^*_{i\rightarrow 1}$ (cf. Eq.~\eqref{eq:spatial_force_transform}). The gravity force acts in negative y-direction; therefore, the spatial gravity force vector is expressed by $f_G =  X^*_{i\rightarrow 1}[0~0~0~0~-m_i g~0]^T$.

The generalized gravity vector is then composed using the sum of the Jacobians of the separate bodies with the corresponding gravity vector (cf. Eq.~\eqref{eq:gravity_force_dynamics}).
\begin{align}
\label{eq:gravity_force_dynamics}
    G(q)&=&\sum^3_{i=1}J_i(q)^Tf_G,~\text{with}\\~X^*_{i\rightarrow 1}&=&\begin{bmatrix}
        R_i & [p_i]_\times R_i\\0 & R_i \end{bmatrix}
        \label{eq:spatial_force_transform}
\end{align}
    
\paragraph{Contact Forces}
As depicted in Fig.~\ref{fig:forces_am}, the contact forces are modeled as a concentrated contact force $F_C$ at the central contact point between the tunnel wall and the pad. Its position is expressed in the local CoSy of the corresponding pads $O_{2/3}$ by $\vec r_\text{contact} = [0~0~r_p]^T$. The contact forces consist of the normal force $F_{n,L/R}$ and the tangential friction forces $F_{x,L/R}$ and $F_{y,L/R}$. These are approximated using Coulomb's Law $F_f=\mu F_n$. Using the enhanced cylindrical contact model proposed in \cite{pereira_enhanced_2015}, the normal force $F_{n,L/R}$
can be expressed by Eq.~\eqref{eq:contact_model} for the case of internal cylindrical contact. Here, $\Delta R$ denotes the difference between the outer radius of the pads and the radius of the tunnel wall, $L$ the length of the pad, $E^*$ the Young composite modulus of the colliding materials \cite{flores_contact_2016}, $\delta$ the penetration depth, $\dot \delta$ the penetration velocity and $\dot \delta^{(-)}$ the velocity at initial contact.
\begin{equation}
    F_{n,L/R}=\frac{aLE^*}{\Delta R}\delta_{L/R}^n\left[1+\frac{3(1-c_e^2)}{4}\frac{\dot \delta_{L/R}}{\dot \delta_{L/R}^{(-)}}\right], 
\label{eq:contact_model}
\end{equation}
where
\begin{equation*}
\begin{aligned}      
a&=(0.965\Delta R+0.0965)\\
~n&=(0.0151\Delta R+1.151)\Delta R^{-5\cdot 10^{-3}}\\
E^*&=\left(\frac{1-\nu_i^2}{E_i}+\frac{1-\nu^2_j}{E_j}\right)^{-1} \label{eq:composit_young_modulus}\\
\delta_{L/R} &= \begin{cases}
    0, & q_{1,2} < \Delta R\\
    q_{1,2}-\Delta R, & q_{1,2} >= \Delta R
\end{cases}
\end{aligned}
\end{equation*}

As the contact forces $F_{C,i}=[F_{x,R/L}~F_{y,R/L}~F_{n,R/L}]^T$ do not act on the CoM of the pads, they exercise a torque on $O_{2/3}$ as shown in Eq.~\eqref{eq:wrench_contact}. The wrench of the contact forces is then transformed using the spatial force transform and extended to generalized coordinates using the Jacobians $J_{2/3}$.
\begin{equation}
    \tau_{C}= \sum^3_{i=2}J_iX^*_{i\rightarrow1}\begin{bmatrix}
        r_{p}\times F_{C,i}\\F_{C,i}
    \end{bmatrix}
    \label{eq:wrench_contact}
\end{equation}

\paragraph{Connecting external wrenches}
To account for the forces and moments resulting from connected modules, wrenches $f_0$ and $f_4$ are added to the model. To form the Jacobian matrices for these wrenches, the spatial force and motion transforms $X^*_{i\rightarrow1}$, and $X_{1\rightarrow i},~i=0/4$, have to be computed. The two corresponding Jacobian matrices $J_{0/4}$ are assembled analogously to $J_1$. The wrench due to connecting modules is then computed by
\begin{equation}
    \tau_{ext}=J_{0}^TX^*_{0\rightarrow1}f_0+J_{4}^TX^*_{4\rightarrow1}f_4.
\end{equation}

\paragraph{Actuator Forces}
The actuator forces do not exert power directly on the reference frame. Therefore, they can be taken into account using $\tau_{act}=[0_6~F_{m,R}~F_{m,L}]^T$.

\subsection{Dynamic Model of the Pushing Module}
\begin{figure}[h]
    \centering
    \includegraphics{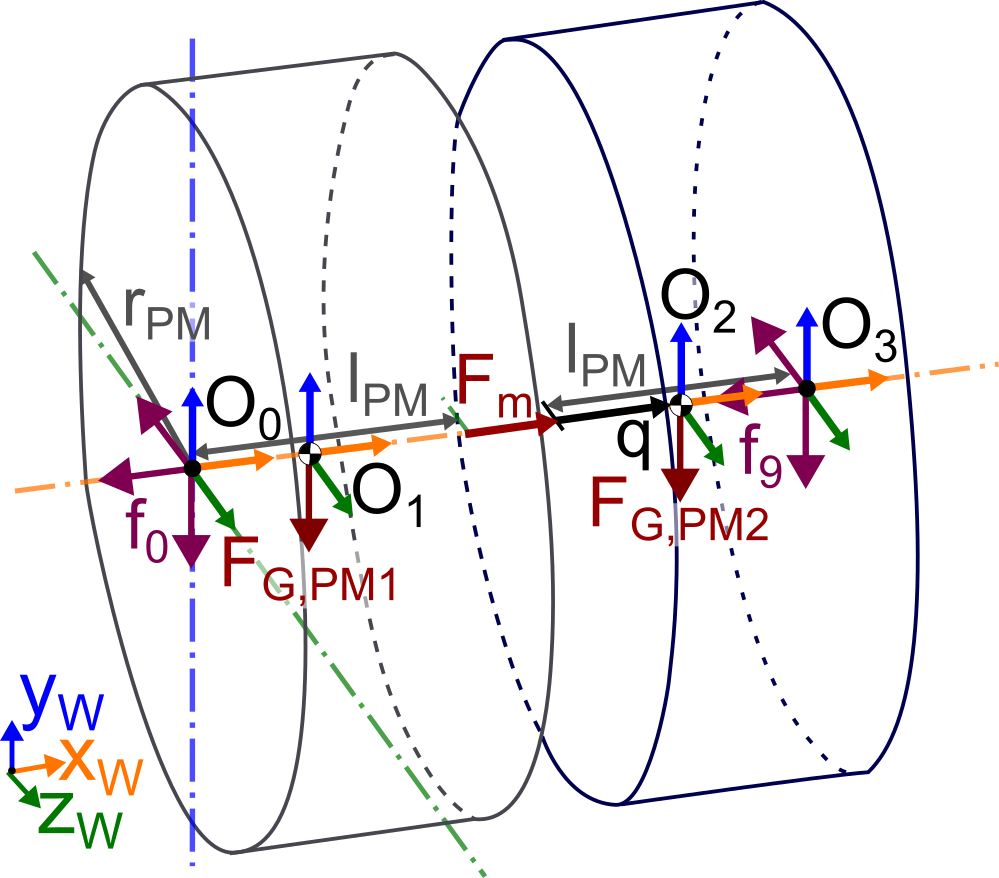}
    \caption{Coordinate systems and forces used for describing the PM}
    \label{fig:forces_pm}
\end{figure}

As described in Section~\ref{sec:description-robot}, the PM utilizes 3 lead-screw systems to push the robot forward. However, as they all share the same single dof $q_{PM}\in[0,15e^{-3}]$\,mm, and must be actuated simultaneously, they are abstracted by one prismatic joint in the model. Similarly to the AM, 2 CoSys act as connection points to other modules and there are CoSys at the CoM of each body. In  total, this leads to 4 CoSys as presented in Fig.~\ref{fig:forces_pm}. All CoSys have the same orientation; thus, the transforms can be described by the translational vectors shown in Tab.~\ref{tab:dh-pm-parameter} and $R=I_3$ as rotational matrix.

\begin{table}[h]
    \centering
    \caption{Positions of the coordinate systems of the PM}
    \begin{tabular}{ccc}
    \hline
    Link$_i$ & Link$_j$& $p_i$\\
    \hline
    0 & 1 & $[l_{PM}/2~0~0]^T$\\
    1 & 2 & $[l_{PM}+q_1~0~0]^T$\\
    2 & 3 & $[l_{PM}/2~0~0]^T$\\
    \hline
    \end{tabular}
    \label{tab:dh-pm-parameter}
\end{table}

\subsubsection{Kinematic relations}
As before, the module is assumed to be free-floating. Therefore, the vector of generalized coordinates is presented by $\dot q = [\omega_0~v_0~\dot q_{PM}]^T$ including the one dof $Q_{PM}\in~[0,20]$\,mm of the prismatic joint. The motion space for the prismatic joint can be described as $S_{PM}=[0~0~0~1~0~0]^T$. The base frame $O_0$ is chosen as reference frame. Using the spatial motion transform described in Eq.~\ref{eq:spatial_motion_transform}, the Jacobian for the first and second body of the PM can be assembled by $J_{0/1} = [X_{0\rightarrow0/1}~0]$ and $J_{2/3}=[X_{1\rightarrow2/3}~S_{PM}]$, respectively. The velocity of the CoSys are described by $v_1=v_0$ and $v_{2/3}=J_{2/3}\dot q$.

\subsubsection{Dynamic relations}
The dynamic equations can be assembled similarly to the ones of the AM. This leads to the following relations:

\begin{equation}
\begin{aligned}
    M(q_{PM}) & =\sum_{i=1}^2J_i^TI_iJ_i
    \label{eq:dynamics_pm}\\
    C(q_{PM},\dot q_{PM}) &  = \frac{1}{2}\left(\frac{\partial M}{\partial q_{PM}}\dot q_{PM}\right)\\
    G & = \sum^2_{i=1}J_i^TX^*_{i\rightarrow1}[0~0~0~0~-m_ig~0^T]\\
    \tau_{ext} & = J_0^TX_{0\rightarrow1}^*f_0+J_3^TX_{3\rightarrow1}^*f_3\\
    \tau_{act} & = [0~F_m]^T
\end{aligned}
\end{equation}

\subsection{Excavation Head Module}
\begin{figure}
    \centering
    \includegraphics{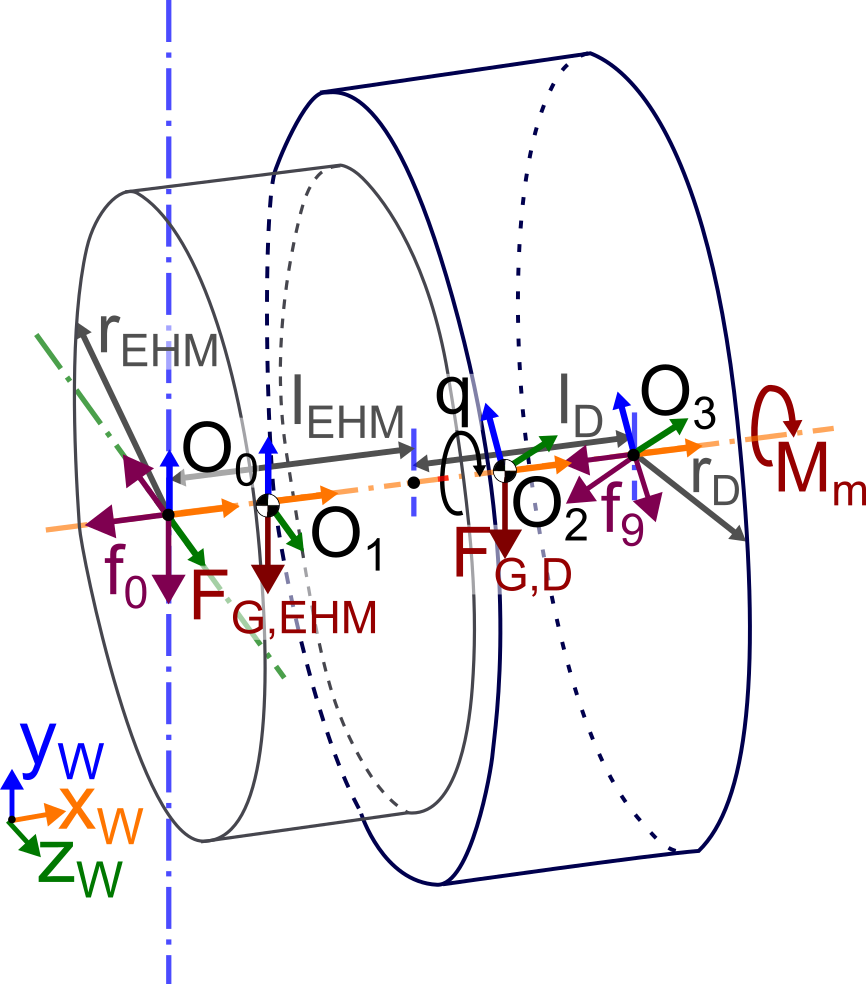}
    \caption{Coordinate system and forces describing the EHM}
    \label{fig:forces_ehm}
\end{figure}

The EHM consists of two bodies connected by a revolute joint as shown in Fig.~\ref{fig:forces_ehm}. Tab.~\ref{tab:dh-ehm-parameter} shows the DH parameters describing the EHM.

\begin{table}[h]
    \centering
    \caption{DH parameter for describing the transforms of the EHM}
    \begin{tabular}{cccccc}
    \hline
    Link$_i$ & Link$_j$& $a_i$ & $\alpha_i$ & $d_i$ & $\theta_i$\\
    \hline
    0 & 1 & $l_{E}/2$ & 0 & 0 & 0\\
    1 & 2 & $(l_{E}+l_{D})/2$ & $q_4$ & 0 & 0\\
    2 & 3 & $l_{D}/2$ & 0 & 0 & 0\\
    \hline
    \end{tabular}
    \label{tab:dh-ehm-parameter}
\end{table}

\subsubsection{Kinematic relations}
The kinematic and dynamic relations are derived using the same principle as those of the AM and PM. The vector of generalized coordinates consists of the 6 dofs for the free-floating base plus the dof $q_D$. However, since the dof comes from a revolute instead of a prismatic joint, its motion space is described by Eq.~\eqref{eq:motion_space_drill}.
\begin{equation}
    S_D = \begin{bmatrix}
        e_x\\e_x\times p_i
    \end{bmatrix}
    \label{eq:motion_space_drill}
\end{equation}

The Jacobian $J_2$ can be assembled using $J_2=[X_{1\rightarrow2}~S_D]$. It is constant as $\dot q$ is expressed with respect to the reference frame and therefore $J_2$ does not include rotational elements.

\subsubsection{Dynamic relations}
\paragraph{Mass and Coriolis Matrix}
The mass and Coriolis matrix is assembled analogously to the PM (cf. Eq.~\eqref{eq:dynamics_pm}), however, as $O_2$ is rotating, its inertia has to be transformed to the reference frame using

\begin{equation}
    I_2^{(1)}=(X^*_{i\rightarrow1})^{-T}I_2^{(2)}(X^*_{i\rightarrow1})^{-1}.
\end{equation}

\paragraph{Gravity Force}
As gravity is always pointing in the negative y-direction and is independent of the rotation of the drill $q_D$, the gravity force of the drill is expressed with respect to reference frame $O_1$. This leads to 
\begin{equation}
    G = [(l_{EHM}+l_D)/2m_Dg~0~0~0~(m_{EHM}+m_D)g~0]^T.
\end{equation}

\paragraph{Drilling Forces and Moments}
As mentioned in Section~\ref{sec:description-robot}, the target excavation force is chosen as 150\,N. Following the work in~\cite{drilling_formulas}, the feed force $F_{f}$ and the cutting moment $M_{\text{exc}}$ can be estimated based on the specific cutting force $k_c$:
\begin{align}
\label{eq:feed_per_revolution}
    F_f\approx 0.25 \cdot D_{drill}\cdot f_n \cdot sin(\kappa_r) \cdot k_c,  \\  
    M_{\text{exc}}  =\frac{f_n\cdot D_{drill}^2\cdot k_c}{8\cdot 10^3}, 
\end{align}
Using the target excavation force as the feed force \(F_f=F_{\text{exc}}\), \( D_{drill} = 190\,mm\) as the drill diameter, $\kappa_r=90^\circ$ as entering angle and the feed per revolution \(f_n = v_f/n \) with $v_f=0.05\,mms^{-1}$ as penetration rate, the specific cutting force and the cutting moment can be calculated depending on the spindle speed $n=v_M$.

This leads to
\begin{equation}
    \tau_{D}=J_3X^*_{3\rightarrow0}[M_\text{exc}~0~0~-F_\text{exc}~0~0]^T
\end{equation}
\section{State-Machine-based Gait Sequence Design}
\label{sec:state-machine}
\begin{figure}
    \centering
    \includegraphics[width=\linewidth]{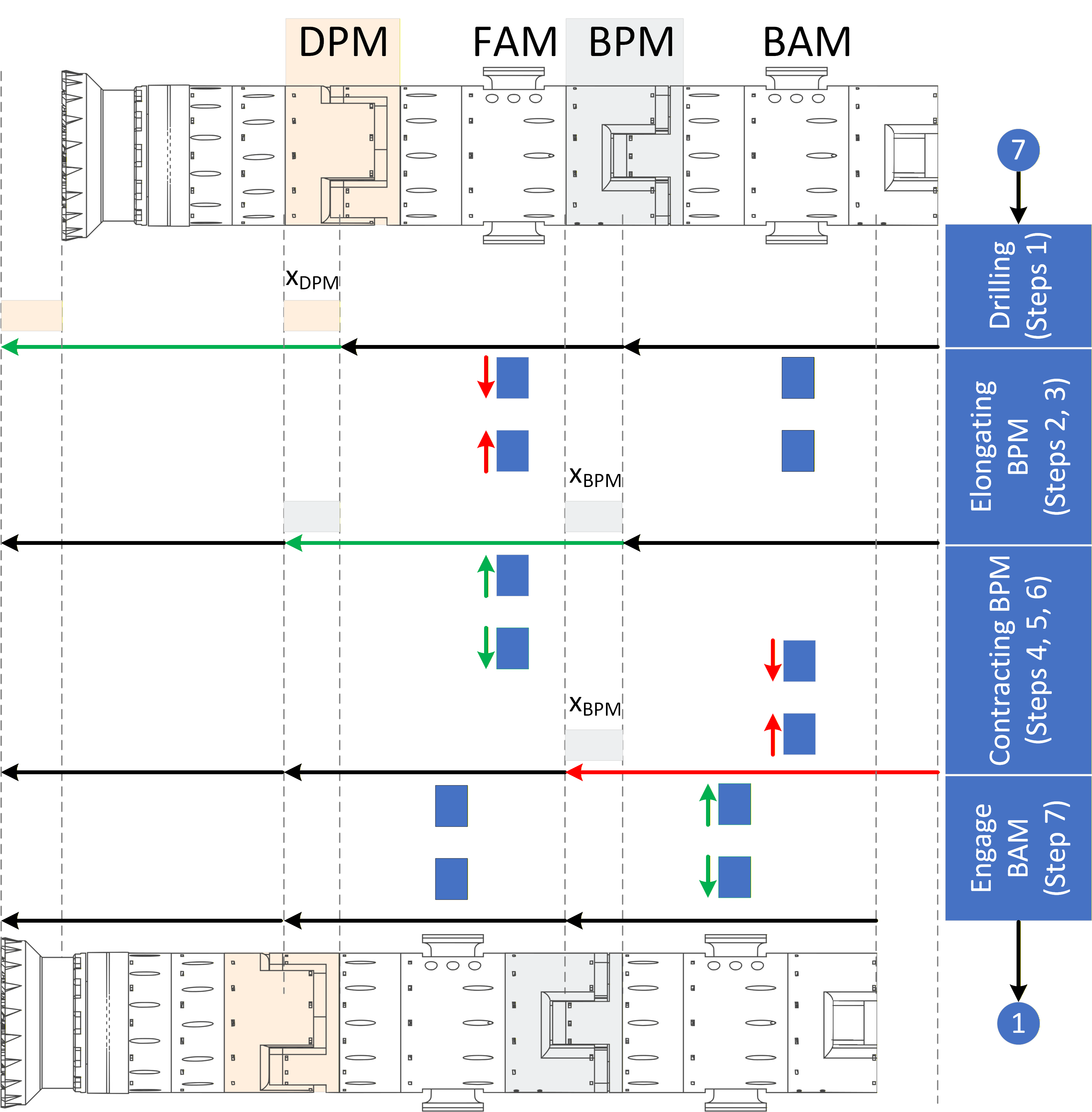}
    \caption{Gait cycle of the robot showing elongation in $x$ direction, where $x_{DPM} = x_{BPM}$. The phases are color-coded: green represents elongation, and red represents contraction.} 
    \label{fig:gait_cycle}
\end{figure}

The robot uses a peristaltic gait cycle as shown in Fig.~\ref{fig:gait_cycle}. Starting with all pads engaged to the tunnel wall following steps have to be executed:
\begin{enumerate}
    \item Drill to the desired drilling depth (DPM activated, drill rotating)
    \item Disengage the front pads
    \item Elongate the BPM to the maximum elongation and simultaneously contract the DPM
    \item Engage the front pads until their normal force reaches the desired value
    \item Disengage the back pads
    \item Contract the BPM until the robot is fully contracted
    \item Engage the back pads until their normal force reaches the desired value
\end{enumerate}

The state machine consists of 5 steps and generates activation signals. It switches when the force on the pads reach the desired anchoring force or the PMs reach there maximal elongation/minimal contraction. Table~\ref{tab:gait_actuators} shows the activation signal generated by the state machine where 1 is used for forward motion, 0 for no actuation, and -1 for backwards motion.

\begin{table}[h]
    \centering
    \caption{Activation signals generated by the centralized state machine}
    \begin{tabular}{lcccc}
    \hline
    State & DPM & FAM & BPM & BAM\\
    \hline
    Drilling & 1 & 1 & 0 & 1\\
   Elongating BPM & -1 & -1 & 1 & 1\\
   Locking FAM & 0 & 1 & 0 & 1\\
   Contracting BPM & 0 & 1 & -1 & -1\\
   Locking BAM & 0 & 1 & 0 & 1\\
   \hline
    \end{tabular}
    \label{tab:gait_actuators}
\end{table}

The robot is designed with a drill diameter of 190\,mm. The modules have a smaller diameter of 160\,mm to leave enough clearance to engage the pads and reduce friction as the robot does not touch the ground during deployment with this configuration. The robot has a length of 0.7\,m and weights around 6\,kg. Due to its design, the pads of the AMs can be extracted to a maximum of 25\,mm while its maximum elongation is 20\,mm. For the first design, a target excavation force $F_\text{exc}$ of 150\,N was chosen.
\section{Controller Design and tuning using dynamic models}
\label{sec:controller}
To verify the EoM of the modules and design custom-tuned controllers, MATLAB/Simulink is used. Again, each module is modeled separately and the velocity of the joints are used as control inputs. A PID-controller is selected as controller for its simple design. In case of the AM, the normal force is used as state feedback and compared to the desired normal force of the pads; for the PM, the position of $O_2$ is selected. Depending on the sign of $\dot q$, the maximal or minimal displacement is used as reference value. In case of the EHM, the motor velocity is used as feedback. The controller generates a velocity command for the motors actuating the joints which is then transformed to the joint velocity $\dot q_{mod}$. Tab.~\ref{tab:controller-gains} shows the selected controller gains for the different modules. The EHM is additionally amplified by a first-order transfer function with a time constant of 0.2. This ensures slow spin-ups, reducing the load on the system during drilling.

\begin{table}[h]
    \centering
    \caption{Controller gains for the different modules}
    \begin{tabular}{cccc}
    \hline
        Gain & AM & PM & EHM\\
        \hline
        $K_P$ & 4 & 100 & 1 \\
        $K_I$ & 1 & 1 & 5\\
        $K_D$ & 0.005 & 0.1 & 0\\
        \hline
    \end{tabular}
    \label{tab:controller-gains}
\end{table}

To check the operability of the designed feedback controller of the robot as well as the plausibility of the derived EoM, tests are run under different operational conditions. Some of these tests will be presented in the following sections.

\subsection{Simulation of the AM}
\begin{figure}[h]
    \centering
    \includegraphics[width=0.9\linewidth]{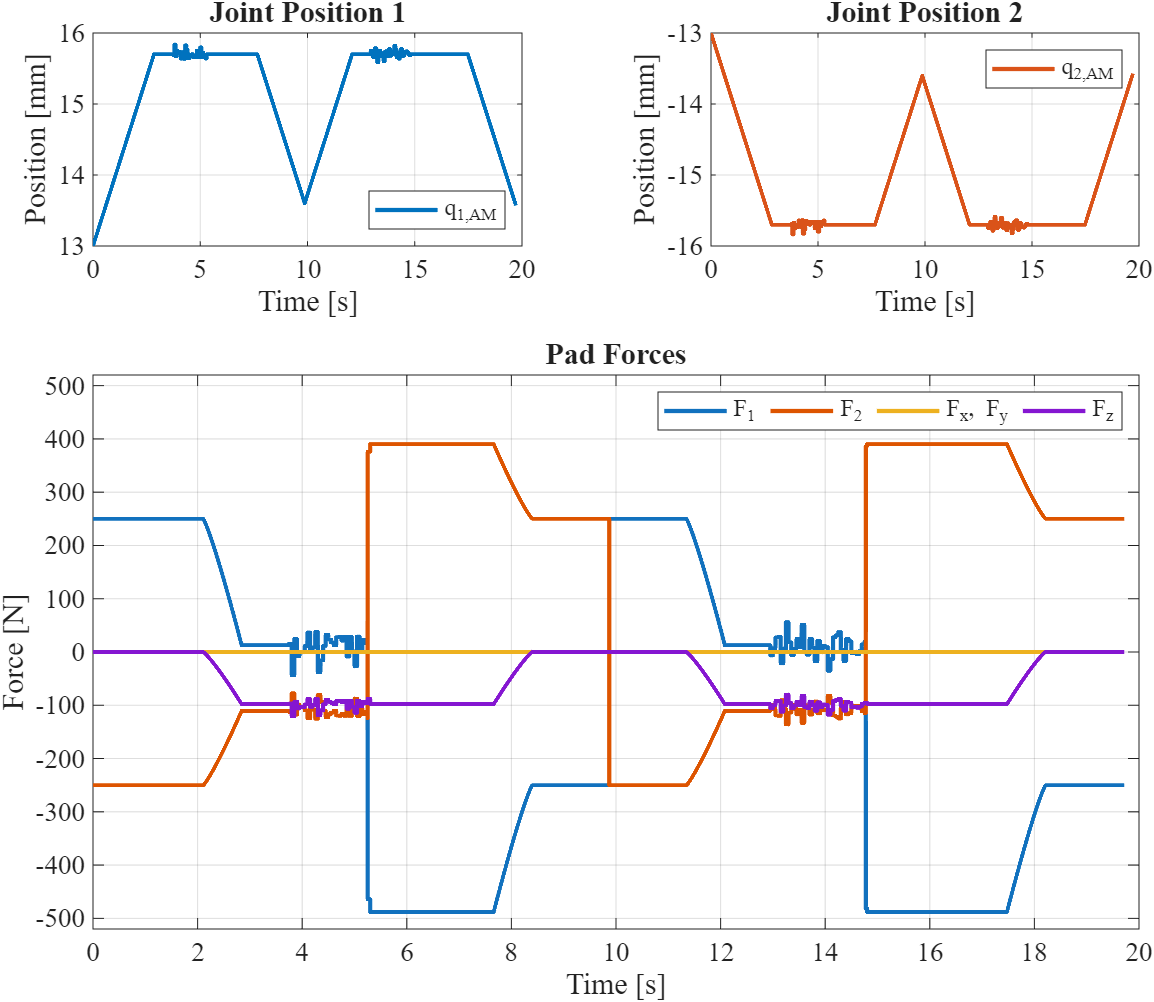}
    \caption{Joint positions and Forces acting on the AM}
    \label{fig:am_simulink}
\end{figure}

The AM is simulated over 20\,s. The model includes a manual switch for selecting the direction of the motor velocity. With that, several anchoring procedures can be tested in the same simulation. Fig.~\ref{fig:am_simulink} shows the joint position and normal forces acting on the pads as well as on the module itself. To show the influence of different soils and test whether the controller can handle these different conditions, the composite young module acting on the left and right pads differ, cf. Eq.~\eqref{eq:contact_model}: $E^*$ amounts $2.188\cdot10^6$ for the first joint and $1.2689\cdot10^6$ for the second joint, respectively. Whenever the pads touch the tunnel walls, here at 15\,mm, the resulting contact forces increase. As can be seen, they rise significantly faster on the pad with higher Young modulus which is plausible as it will deform less than the one with lower Young modulus. Additionally, the module itself is pushed towards the second joint and reduces the force necessary to settle in a fixed position. After settling, a band-limited white noise with sampling time $5\cdot10^{-2}$ and limit of $1\cdot10^{-10}$\,Hz is added to simulate disturbances during drilling operations. Even with these small disturbances, force disturbances of almost 50\,N can be observed showing that even small disturbances have a big impact on the overall dynamics. When the motor direction is switched and the pads disengage, the normal forces decrease until solely the motor force is acting on the joints. Also, the load on the module decreases to 0 again. The second anchoring procedure shows a trend similar to that of the first. During the complete simulation the velocity as well as position of the CoM did not change showing that the development of the controller can be assumed successful.

\subsection{Simulation of the PM}
\begin{figure}
    \centering
    \includegraphics[width=\linewidth]{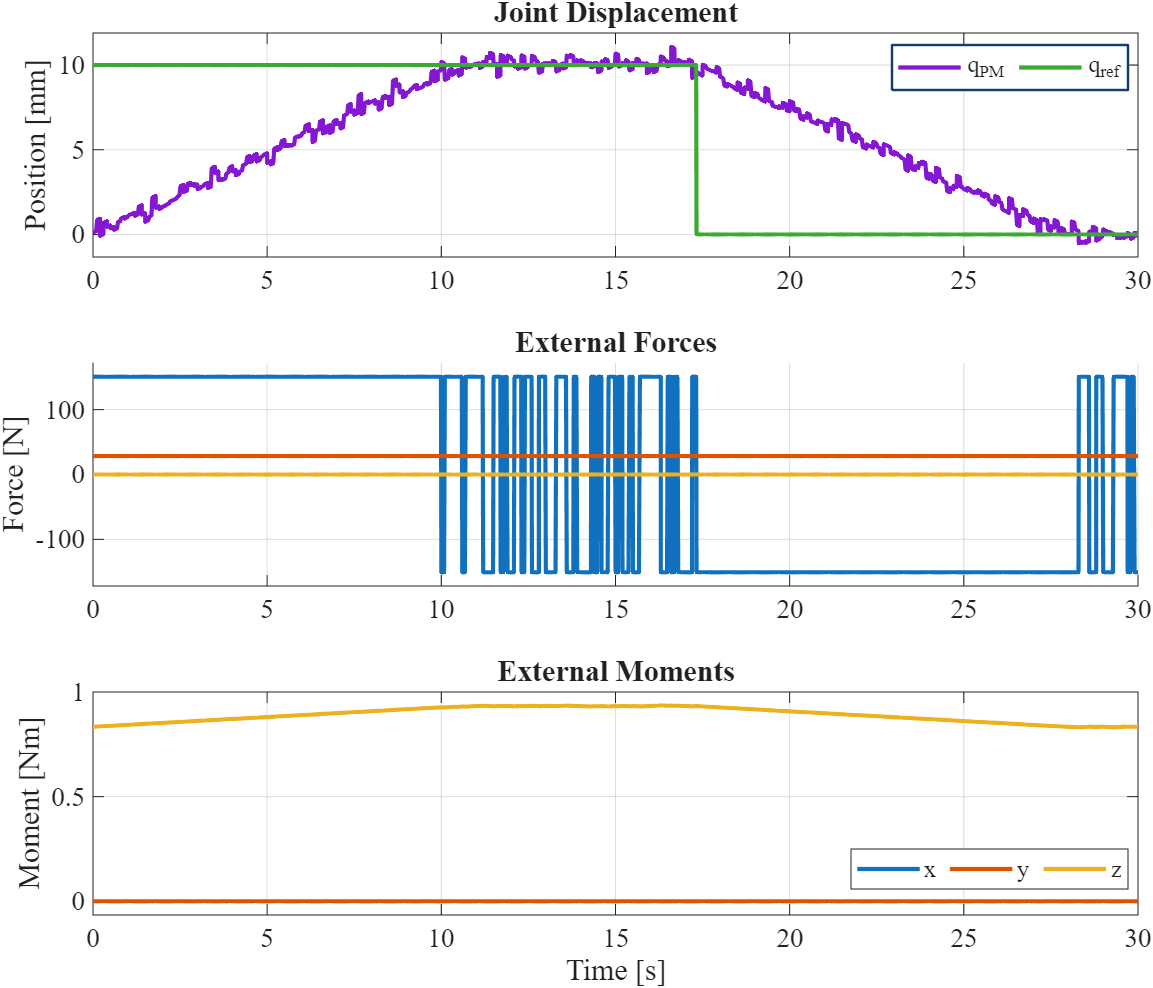}
    \setlength{\abovecaptionskip}{-13pt}
    \caption{Simulation of the dynamic model of the PM}
    \label{fig:pm_simulink}
\end{figure}

As the PM is usually connected to the AM and all occurring moments and forces are countered by its pads, occurring wrenches are countered at $O_0$ (cf. Fig.~\ref{fig:forces_pm}. Also this model contains a manual switch which changes the reference value to the minimal and maximal displacement of the joint $q_{PM}$. Fig.~\ref{fig:pm_simulink} shows a simulation of the dynamics of the PM. As for the AM, white-noise is added to the model to account for drilling disturbances. By choosing $0_0$ as reference frame, a static moment in y-direction can be observed which is increased when elongating the module. During movement, $F_x$ counters the actuator forces $F_m$. When the controller reaches it settling point, the white-noise starts to have an impact on the module dynamics: To counter the small displacements of $q_{PM}$, the controller has to activate the actuator again and its direction switches rapidly. Nonetheless, the controller manages to control the disturbed position showing its ability to be used on the real robot.

\subsection{Simulation of the EHM}
\begin{figure}
    \centering
    \includegraphics[width=\linewidth]{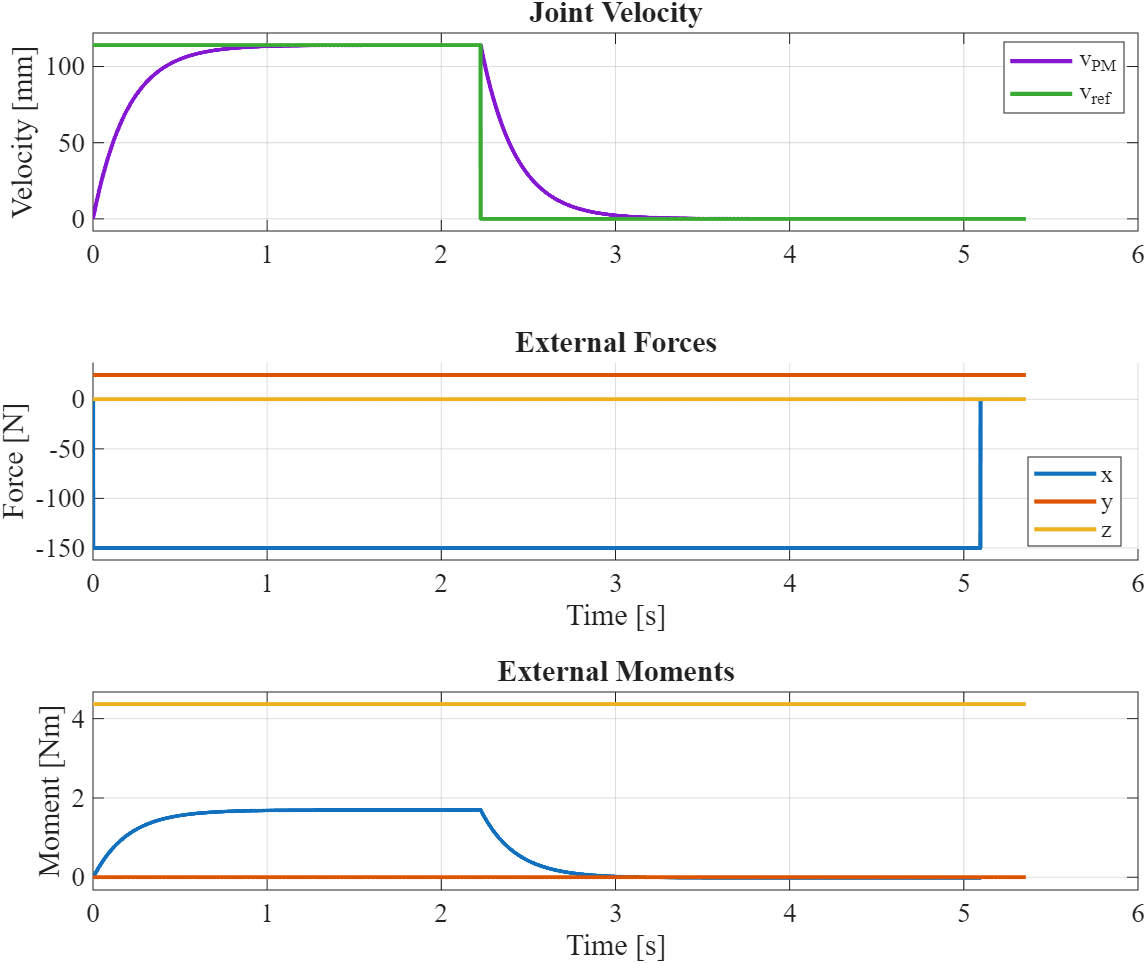}
    \setlength{\abovecaptionskip}{-13pt}
    \caption{Simulation of the dynamic model of the EHM}
    \label{fig:ehm_simulink}
\end{figure}

Also in this case, the occurring forces and moments are usually countered by the pads of the AM and therefore wrench $f_0$ is added to the system (cf Fig.~\ref{fig:forces_ehm}). Whenever the motor spins, the drilling force and moment are acting on the EHM. As can be seen in Fig.~\ref{fig:ehm_simulink}, by adding the first-order transfer function, the drilling moment is influenced and there is no abrupt rise of moment due to drilling.

\section{Unity Simulation of the subsurface locomotion of the robot}
\begin{figure*}
    \centering
    \includegraphics[width=\linewidth]{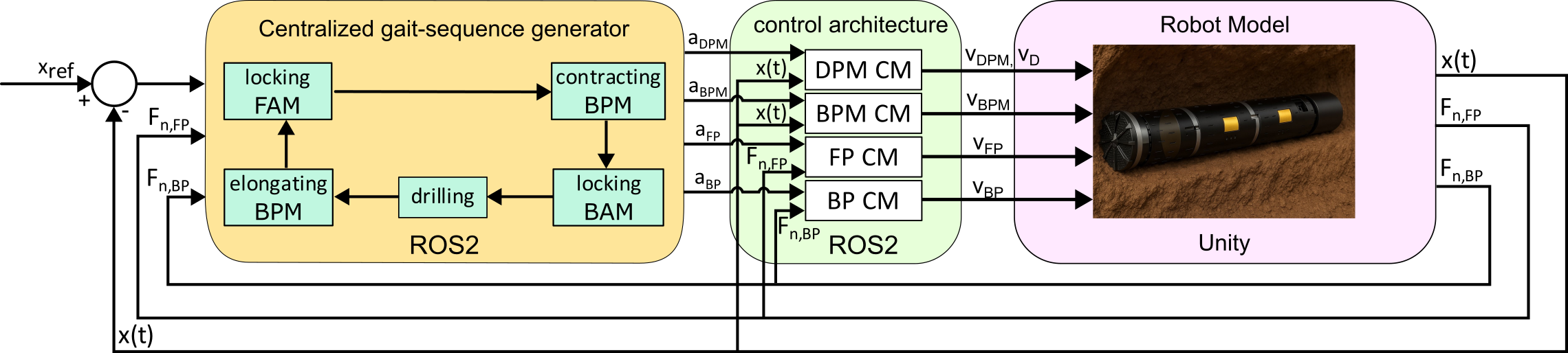}
    \caption{Block diagram of the Unity simulation}
    \label{fig:block_diagram_sim}
\end{figure*}

For the simulation of the complete robot Unity \cite{unity} and Robot Operating Systems (ROS) are used. The activation signals generated by the centralized state-machine described in Section~\ref{sec:state-machine} is used in the control module (CM) of each module to select the reference value and generate velocity commands using the controllers developed in Section~\ref{sec:controller}. The complete block diagram of the simulation is shown in Fig.~\ref{fig:block_diagram_sim}.

\begin{figure}[h]
    \centering
    \includegraphics[width=0.98\linewidth]{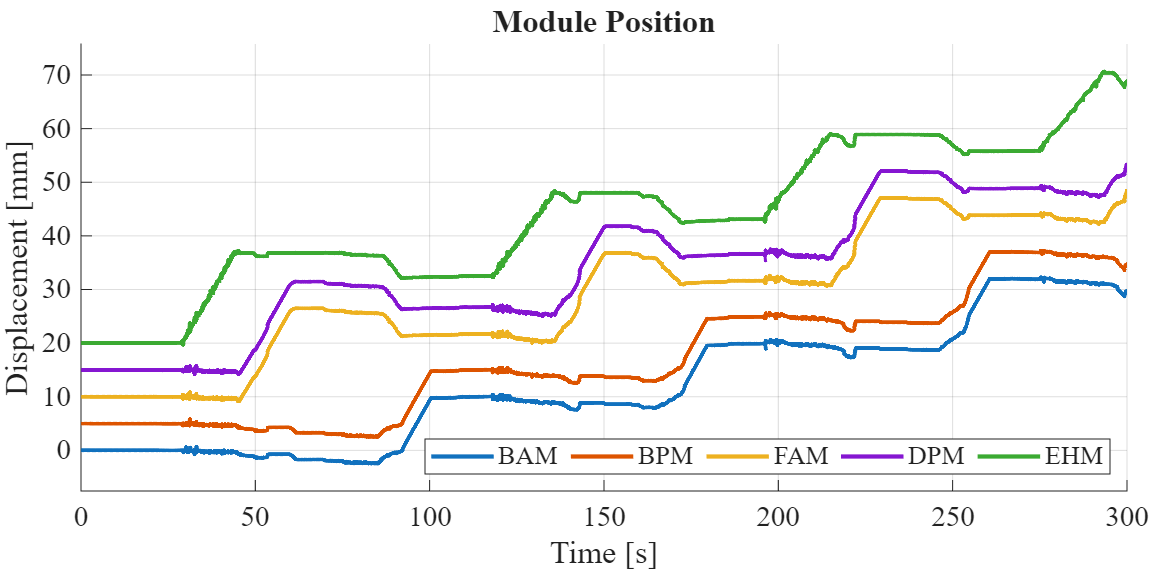}
    \setlength{\abovecaptionskip}{-3pt}
    \caption{Module position in the Unity simulation}
    \label{fig:module_positions_unity}
\end{figure}

 To simulate interactions between the soil and the drill, the drill pushes a heavy mass forward. Due to its rotation and the non-linearity of the contact, this leads to disturbances which have to be countered by the developed controllers. The simulation is run over 300\,s to show the propagation during several drilling cycles. During that time, the robot completes 3 full gait cycles. The module positions are plotted in Fig.~\ref{fig:module_positions_unity}. To emphasize the propagation itself, the distances between the modules were replaced by a fixed distance of 5\,mm. In total, the robot moved 30\,mm forward. In theory, 60\,mm could be achieved. The most significant influence for this reduction is a slippage when the BPM contracts. So far, it is unclear, why this slippage occurs as it is negligible when the DPM contracts, however a possible explanation is that the chosen anchoring force of 250\,N is not sufficient to handle the increased load occurring due to the lever arm of the BPM. Additionally to the slippage, the state machine and the controller include safety thresholds which ensure that the maximal possible elongations will not be reached to reduce loads on the material of the robot.
 
\begin{figure}[h]
    \centering
    \includegraphics[width=0.98\linewidth]{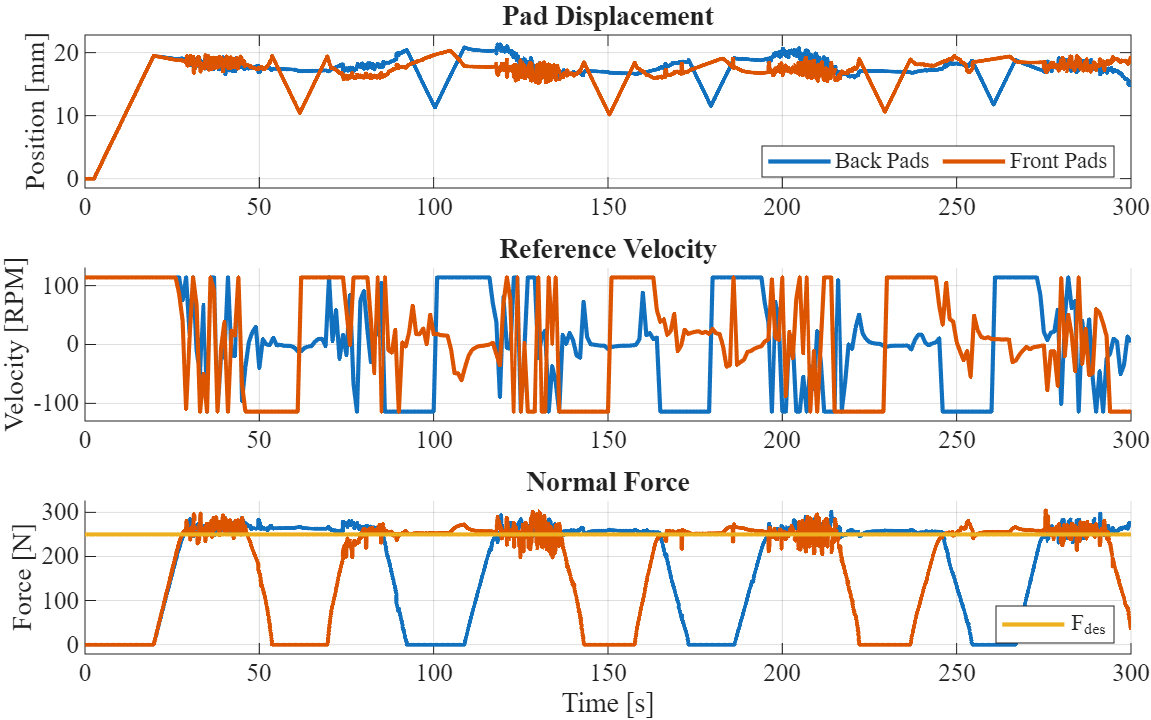}
    \setlength{\abovecaptionskip}{-3pt}
    \caption{Pad displacement, reference velocity and normal forces acting on the AMs during simulation}
    \label{fig:pad_simulation}
\end{figure}

In Fig.~\ref{fig:pad_simulation}, the joint displacement $q_{BAM}$ and $q_{FAM}$, the reference velocity $\dot q_{BAM}$ and $\dot q_{FAM}$ generated by the controller and the normal forces acting on the pads are plotted. As can be seen, at least one AM is always engaged to the tunnel wall. This ensures that the robot body does not touch the tunnel during operation. Whenever all pads are engaged to the tunnel wall, the robot starts drilling. The oscillations observed in the forces stem from the described drilling interaction. These are countered by the controller; however, it takes a high control effort to do so.

\begin{figure}[h]
    \centering
    \includegraphics[width=0.98\linewidth]{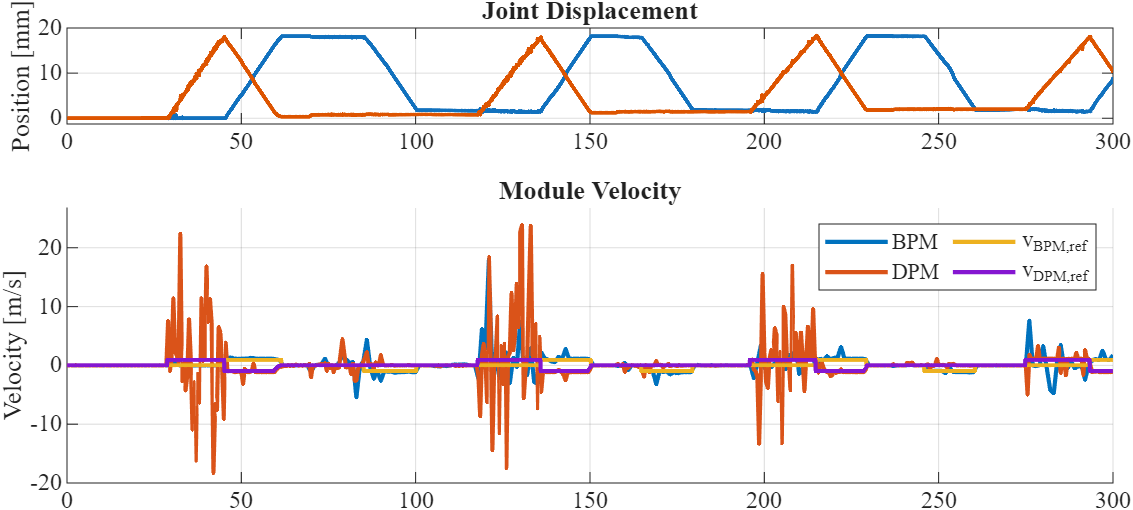}
    \setlength{\abovecaptionskip}{-3pt}
    \caption{Joint displacement and module velocity of the PMs}
    \label{fig:pm_simulation}
\end{figure}

The joint displacement and the module velocity of the PMs are presented in Fig.~\ref{fig:pm_simulation}. Also in this figure, the disturbances induced by the drilling interaction are clearly visible. Starting from the second drilling phase, the oscillations also affect the BPM. This can be explained by the safety threshold. As the BPM does not fully contract to 0\,mm displacement, the controller holds this reference value instead.

\section{Conclusion}
This paper presented the modeling and simulation of a modular bio-inspired drilling robot aimed for deep exploration. The dynamics of the separate modules and verified using several Simulink simulations. Additionally, PID-controllers were tuned to control the anchoring force of the AMs, the position of the PMs and the speed of the drill. A physics-based simulation using Unity and ROS was developed. The controller generates activation signals of a peristaltic gait based on a centralized state-machine and applies the developed controller on these signals. Lastly, effective forward propagation was demonstrated in simulation in the presence of disturbances occurring in the drilling operation. Future work will focus on updating the robot and simulation to achieve 3-dimensional movement and developing a trajectory tracking controller enabling contour following while considering the gait movement cycle.

\bibliographystyle{IEEEtran}
\bibliography{bibliography}

\end{document}